\documentclass{article}

\usepackage[preprint]{neurips_2023}

\usepackage[utf8]{inputenc} 
\usepackage[T1]{fontenc}    
\usepackage{hyperref}       
\usepackage{url}            
\usepackage{booktabs}       
\usepackage{amsfonts}       
\usepackage{nicefrac}       
\usepackage{microtype}      
\usepackage{xcolor}         

\usepackage{fancyhdr}
\usepackage{tabu}
\usepackage{tabularx}

\usepackage{makecell}
\usepackage{amsmath,amssymb,amsfonts}
\usepackage{algorithmic}
\usepackage{algorithm}
\usepackage{graphicx}
\usepackage{textcomp}
\usepackage{xcolor}
\usepackage{enumitem}
\usepackage{subfigure}
\usepackage{soul}

\newtheorem{mdefinition}{Definition}

\newcommand{\para}[1]{{\vspace{3pt} \bf \noindent #1 \hspace{0pt}}}
\usepackage{enumitem}

\title{EPR-GAIL: An EPR-Enhanced Hierarchical Imitation Learning Framework to Simulate Complex User Consumption Behaviors}

\author{%
  Tao Feng \\
  Department of Electronic Engineering\\
  Tsinghua University\\
  Beijing, China \\
  \And
  Yunke Zhang \\
  Department of Electronic Engineering\\
  Tsinghua University\\
  Beijing, China \\
  \And
  Huandong Wang \\
  Department of Electronic Engineering\\
  Tsinghua University\\
  Beijing, China \\
  \And
  Yong Li \\
  Department of Electronic Engineering\\
  Tsinghua University\\
  Beijing, China
}

\begin{document}

\maketitle

\begin{abstract}
  User consumption behavior data, which records individuals' online spending history at various types of stores, has been widely used in various applications, such as store recommendation, site selection, and sale forecasting.
However, its high worth is limited due to deficiencies in data comprehensiveness and changes of application scenarios.
Thus, generating high-quality sequential consumption data by simulating complex user consumption behaviors is of great importance to real-world applications.
Two branches of existing sequence generation methods are both limited in quality. \emph{Model-based methods} with simplified assumptions fail to model the complex decision process of user consumption, while \emph{data-driven methods} that emulate real-world data are prone to noises, unobserved behaviors, and dynamic decision space. 
In this work, we propose to enhance the fidelity and trustworthiness of the \emph{data-driven} Generative Adversarial Imitation Learning (GAIL) method by blending it with the Exploration and Preferential Return (EPR)~\emph{model}.
The core idea of our EPR-GAIL framework is to model user consumption behaviors as a complex EPR decision process, which consists of purchase, exploration, and preference decisions. Specifically, we design the hierarchical policy function in the generator as a realization of the EPR decision process and employ the probability distributions of the EPR model to guide the reward function in the discriminator.
Extensive experiments on two real-world datasets of user consumption behaviors on an online platform demonstrate that the EPR-GAIL framework outperforms the best state-of-the-art baseline by over 19\% in terms of data fidelity. Furthermore, the generated consumption behavior data can improve the performance of sale prediction and location recommendation by up to 35.29\% and 11.19\%, respectively, validating its advantage for practical applications.
\end{abstract}

\section{Introduction}






Consumption behavior data records consumers' spatio-temporal buying activities at various providers.
Understanding the underlying and decisive patterns of user consumption behaviors
is of great importance to a variety of applications, such as store recommendation~\cite{kowatsch2010store}, sales prediction~\cite{kaneko2016deep}, and site selection~\cite{guo2018citytransfer,liu2019deepstore}.
Particularly, in the burgeoning e-commerce era, online platforms like Meituan~\cite{li2022automatically,ping2021user} and Meta~\cite{yaacoub2016effect} profoundly rely on user consumption behavior data to predict potential purchases and make marketing decisions for substantial commercial revenues.
However, the availability and applicability of user consumption behavior data are often restricted by the limitation of data comprehensiveness, \emph{e.g.}, missing consumption records or covering limited categories of stores.
Meanwhile, directly replicating real-world data would make it hard to model counterfactual scenarios~\cite{yuan2022activity}. As a result, the realistic simulation of user consumption behaviors to produce artificial purchase sequences data with fidelity and utility still remains a valuable but challenging problem.

In this paper, we are devoted to generating massive consumption behavior sequences by integrally simulating individuals' decision processes of \emph{whether}, \emph{where} and \emph{when} to purchase, while preserving the original characteristics of real-world data.
Existing solutions of simulating user consumption behavior data can be categorized into two branches, \emph{i.e.}, \emph{model-based methods} and \emph{data-driven methods}.
The model-based methods generate consumption sequences explicitly with mathematical and statistical models derived from economical, sociological, or psychological theories~\cite{engel1986consumer, ajzen1996social, said2001multi}.
In particular, such methods often assume that user consumption behaviors can be represented by a decision process that only features a limited number of factors (\emph{e.g.}, consumer's demand and store preference), which are not fully capable of simulating complex decision processes with complex factors based on spatial and temporal dependencies~\cite{ajzen1996social}, especially for consumption behaviors on online platforms.
On the other hand, the recent advances of generative adversarial networks~\cite{yu2017seqgan, chen2016infogan} and imitation learning~\cite{torabi2018behavioral, ng2000algorithms} shed light on directly fitting real-world consumption behavior data through data-driven approaches.
Despite that data-driven approaches are empowered by neural networks and have succeeded in imitating real-world data with better performance,
still they cannot fully capture some hidden and decisive patterns~\cite{yuan2022activity}
due to the lack of prior knowledge of real-world consumption behaviors.

In this paper, we propose a novel framework by integrating the advantages of \emph{model-based methods} and \emph{data-driven methods}, to achieve high-quality user consumption behavior simulation.
Our core idea is to leverage a specially designed Exploration and Preferential Return (EPR) model~\cite{song2010modelling}
as the guidance of users' consumption decision process in a \emph{data-driven} Generative Adversarial Imitation Learning (GAIL) framework~\cite{ho2016generative}.
We customize the EPR model to depict the spatial choices of individuals as a two-stage decision process and further capture the characteristic of user consumption behaviors.
In particular, the two-decision process explicitly explains two kinds of key decisions by users,
\emph{i.e.}, the decision on whether to explore new stores or return to visited places, and the decision on which location to visit.
We implement such a decision process and its statistical characteristics into both the generator and the discriminator of a customized GAIL framework.
Specifically, we design a hierarchical policy function that consists of a purchase agent, an exploration agent, and a preference agent in the generator as the realization of the EPR model to generate consumption decisions based on various factors.
Moreover, we blend the probability distribution of location choices of the EPR model into the reward function,
with the goal to inspire the discriminator to distinguish user consumption behaviors by considering the overall distribution of  different consumption choices in the generated sequences.
Our contributions can be summarized as follows: (1) We propose a novel EPR-enhanced hierarchical generative imitation learning framework (EPR-GAIL), which exploits the advantages of \emph{model-based methods} and \emph{data-driven methods} to  efficiently generate sequential consumption data of high quality;
(2) We propose to utilize the prior knowledge of the exploration and preferential return model by modeling the EPR decision process in the generator; (3) Extensive experiments on two real-world consumption datasets demonstrate that the proposed EPR-GAIL outperforms the state-of-the-art baseline approaches by reducing the Jensen-Shannon divergence metrics by more than 19.04\%. Furthermore, we apply the generated consumption sequence data in sale prediction and store recommendation tasks, and our generated data  improves the mean average percentage errors and overall accuracy by over 35.29\% and 11.19\%, respectively.

\section{Problem Formulation}
We consider a general e-commerce scenario,
where customers use the mobile App to access the online platform and explore different stores for shopping and consumption.
The consumption history of each user is composed of a series of spatio-temporal records.
To pave the way for modeling the spatio-temporal sequences from the consumption datasets with consistency and under different resolutions, we divide time into fixed-length slots and utilize identifiers to represent different online stores.
We denote $\mathcal{U}$, $\Omega$ and $\mathcal{T}$ as the set of users, the set of stores and the set of time slots, respectively. 
Then, for each user $u\in\mathcal{U}$,
his/her consumption history can be represented by $T_u=\{(t_1,l_1,f_1),(t_2,l_2,f_2),...,(t_N,l_N,f_N)\}$.
Here, $l_i$ is the identifier of the stores with consumption history,
$t_i$ is the corresponding timestamp, $N$ is the length of consumption sequence,
and $f_i$ is the feature set of online store $l_i$.
Note that $f_i$ includes the category of stores $c_{i}$, the average price of stores $g_{i}$, the historical visits matrix of stores $v_{i}$, and the distance matrix between individual users and stores $d_{i}$.
In addition, the gap between two consecutive time slots is identical and denoted by $\Delta$,
i.e., $\Delta=t_{2}-t_{1}=t_{3}-t_{2}=...=t_{N}-t_{N-1}$.
In particular, we formulate the non-consumption behavior as a special type of `store' and use identifier `0' to denote it in users' historical consumption sequences.
At last, we use $T^{g}$ and $T^{r}$ to represent the generated consumption sequence data and real-world user consumption sequence data, respectively. Based on the above notations, the user consumption sequence synthesizing problem can be defined as follows.

\begin{mdefinition}[Online Consumption Sequence Generation Problem]\label{def:problem}
Given a set of users $\mathcal{U}$ and their historical consumption sequences $\{T_u\}_{u\in\mathcal{U}}$,
the goal is to train a robust generator to simulate synthetic user consumption sequences.
The synthetic sequences are expected to be differential from the original data while preserving high-level fidelity and utility.
\end{mdefinition}

\section{Method}\label{sec:gail}

\begin{figure*}[t]
\centering
\includegraphics*[width=0.99\textwidth]{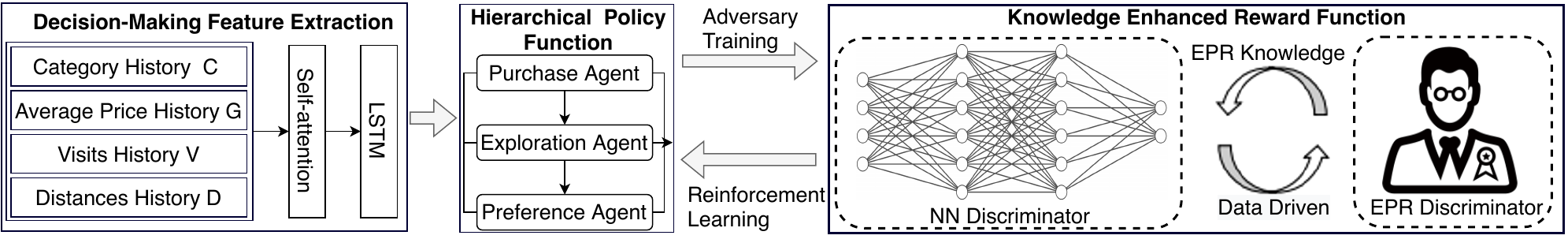}
\caption{The overall illustration of the EPR-GAIL framework.}~\label{fig:framework}
\vspace{-3pt}
\end{figure*}

The design purpose of EPR-GAIL is to combine the advantages of both model-based methods and data driven methods, thus to achieve reliable and robust simulation on user consumption behaviors. Figure~\ref{fig:framework} illustrates the solution framework of the proposed EPR-GAIL.
The architecture of EPR-GAIL has three interconnected components,
including a decision-making feature extraction module,
a hierarchical policy function, and a knowledge-enhanced reward function.

\subsection{Decision-Making Feature Extraction}\label{sec:policy}
To capture complex sequential transition regularities in user consumption sequences,
we design a decision-making feature extraction module with both a self-attention layer~\cite{vaswani2017attention} and an LSTM layer.
As shown in Figure~\ref{fig:feature}, we fuse different types of heterogeneous features to feature embeddings $x^{c}$ through the self-attention layer, which contains the category information $C_{i}$, the average price information $G_{i}$, the visit information $V_{i}$ and the historical user-store distance $D_{i}$. Then the LSTM  extracts the sequential information $x^{l}$ from the feature embedding trajectories $(x_1^{c},x_2^{c},...,x_i^{c})$ of $i$-th time interval, which serves as the feature representation for the decision-making process of the hierarchical policy function. The joint usage of two networks can help us better extract feature history, which has been studied in many works \cite{nagrani2021attention,bui2014model}. The details of  this part are introduced in Appendix \ref{sec:Feature Extraction}.

\begin{figure} [t!]
\begin{center}
\vspace{-2mm}
\includegraphics*[width=0.69\textwidth]{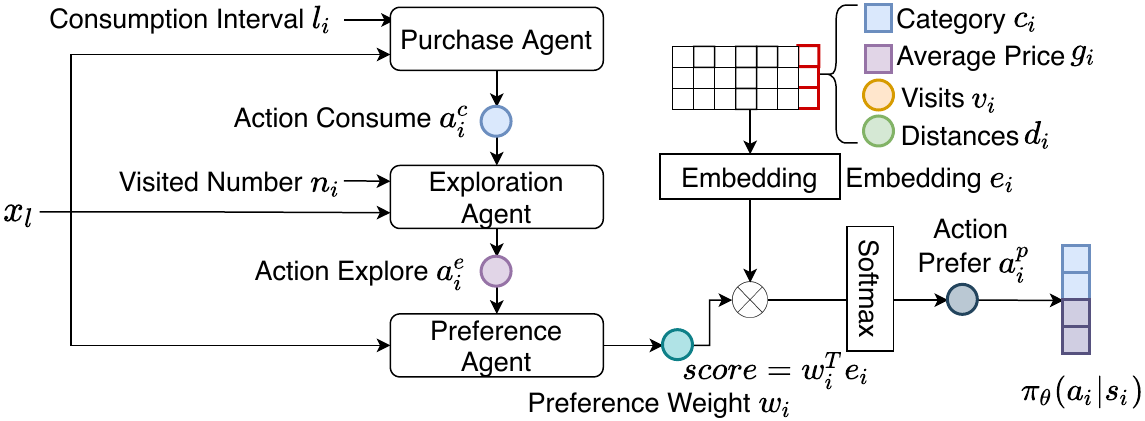}
\end{center}
\caption{Hierarchical Policy Function.} \label{fig:HRL}
\vspace{-3pt}
\end{figure}

\subsection{Hierarchical Policy Function}
Some psychological studies have shown that there is a hierarchical structure in human's mind
during the psychological decision-making process~\cite{chen2020understanding,su2020consumer}.
In correspondence, during the decision-making process of users choosing stores for consumption, their psychological choices also form a hierarchical structure.
On the top of such structure, users tend to first think about whether to purchase or not~\cite{chen2020understanding,su2020consumer}.
If they are to purchase, they will then think about whether to purchase at stores they have consumed or new stores that they have never visited online before.
Finally, they will decide which store to purchase.
As shown in Figure~\ref{fig:HRL},
inspired by the above psychological flow,
we model users' complex consumption behaviors through a three-layer hierarchical policy function.
For each layer, we model the decision process as a Markov Decision Process (MDP).
In the following, we will introduce the above three layers in detail.

\para{The Purchase Agent.}
This agent is at the upper level of the hierarchical policy function and it models whether the user purchases or not.
The detail of MDP setting is as follows.
\begin{itemize}
    \item
    \textbf{State:} 
    At each time step $i$, the state of purchase agent $s_{i}^{c}$ includes the feature representation vector $x_{i}^{l}$ and the purchase interval $l_{i}$~\cite{jiang2016timegeo}. Note that $x_{i}^{l}$ is embedded by the decision-making feature extraction model and $l_{i}$ contains information of users' purchase patterns as shown in Figure~\ref{fig:HRL}. Therefore, we can obtain $s_{i}^{c}=(x_{i}^{l},l_{i})$.
    
    \item 
    \textbf{Action:}
    The action of the purchase agent $a_{i}^{c}$ indicates whether the user purchases at time step $i$. For example, if a user chooses to purchase at time step $i$, then $a_{i}^{c}=1$.
\end{itemize}

\para{Exploration Agent.}
This agent is at middle level of hierarchical policy function and it models whether the user decides to explore a store he/she has never purchased before or return to a store he/she has purchased before. The detail of MDP setting is as follows.
\begin{itemize}
    \item
    \textbf{State:} 
    Choosing to explore or return is based on whether to purchase.
    Therefore, at each time step $i$, the state of explore agent $s_{i}^{e}$ includes the feature representation vector $x_{i}^{l}$ embedded by decision-making feature extraction model, visited store number $n_{i}$ which contains information about user explore  patterns~\cite{jiang2016timegeo} and the action purchase $a_{i}^{c}$ given by purchase agent as shown in Figure~\ref{fig:HRL}. Therefore, we can obtain $s_{i}^{e}=(x_{i}^{l},n_{i},a_{i}^{c})$.
  
    \item 
    \textbf{Action:}
    The action of purchase agent $a_{i}^{e}$ indicates whether the user explores a store he/she has never purchased before at time step $i$. For example, if the user explores a store he/she has never purchased before at time step $i$, then $a_{i}^{e}=1$.
\end{itemize}

\para{Preference Agent.}
This agent is at lower level of hierarchical policy function and it models which store the user decides to purchase. The detail of MDP setting is as follows.
\begin{itemize}
    \item
    \textbf{State:} 
    Choosing which store to purchase is based on the decision of whether to purchase and whether to explore. Therefore, at each time step $i$, the state of preference agent $s_{i}^{p}$ includes the feature representation vector $x_{i}^{l}$ embedded by the decision-making feature extraction model, action purchase $a_{i}^{c}$ given by the purchase agent and action explore $a_{i}^{e}$ given by the explore agent, as shown in Figure~\ref{fig:HRL}. Therefore, we can obtain $s_{i}^{p}=(x_{i}^{l},a_{i}^{c},a_{i}^{e})$.
  
    \item 
    \textbf{Action:}
    If we directly set the action of the preference agent as selecting a specific store, then its action space must be fixed. However, in the real-world scenario, the number of stores that users can choose to purchase will change when new stores go online or old stores close. Therefore, such method is not practical. To address this dilemma, we draw on a point of view from purchaser psychology~\cite{chen2020understanding,su2020consumer}: customers' different choices on stores actually reflect different preferences for stores' features. Inspired by this, we have specially designed a preference agent as shown in Figure~\ref{fig:HRL}. The preference agent first embeds the state $s_{i}^{p}$ and outputs preference weight of store features $w_{i}\in \mathbb{R}^{m \times d }$ by 
    \begin{equation}\label{equ:pate}
    w_{i}=Relu(s_{i}^{p}W_{p}),
    \end{equation}
    where $W_{p}$ is the learnable parameter, $m$ is the number of users and $d$ is the dim of embedding.
    Then we convert the store feature matrix into a dense feature embedding $e_{i}\in \mathbb{R}^{d\times n}$ through a store feature embedding layer by 
    \begin{equation}\label{equ:pate}
    e_{i}=Relu([c_{i},g_{i},v_{i},d_{i}]W_{e}),
    \end{equation}
    where $W_{e}$ is the learnable parameter, $n$ is the number of stores, $c_{i}$ is the categories of stores, $g_{i}$ is the average price of stores, $v_{i}$ is historical visits matrix of stores, $d_{i}$ is distance matrix between user and stores. 
    Inspired by the design of attention mechanism for supporting model's inputs with variable lengths~\cite{vaswani2017attention}, we regard the $w_{i}$ and $e_{i}$ as key and query to apply inner product for $w_{i}$ and $e_{i}$ and obtain the preference $Score \in \mathbb{R}^{m\times n}$ of all stores,
    \begin{equation}\label{equ:pate}
    Score=w_{i}^{T}e_{i}.
    \end{equation}
    With the above design, our model works smoothly when new stores are online or old stores close. In addition, based on the purchase action and exploration action, we add a mask operation for the $Score$. For example, if the explore agent choose to explore, then we mask  the scores corresponding to customers with purchase histories in the corresponding stores.
    Finally, we obtain the action preference $a_{i}^{p}$ through a soft-max layer,
    \begin{equation}\label{equ:pate}
    a_{i}^{p}=\mathrm{softmax}(Score). 
    \end{equation}
    The action preference gives the probability of choosing each store to purchase.
\end{itemize}

In summary, the hierarchical policy function is a decision strategy with state $s_{i}=(s_{i}^{c},s_{i}^{e},s_{i}^{p})$ as input and $a_{i}=(a_{i}^{c},a_{i}^{e},a_{i}^{p})$ as output, which is parameterized by $\theta$ and denoted by $\pi_\theta(a_{i}|s_{i})$.

\begin{figure} [t!]
\begin{center}
\includegraphics*[width=0.62\textwidth]{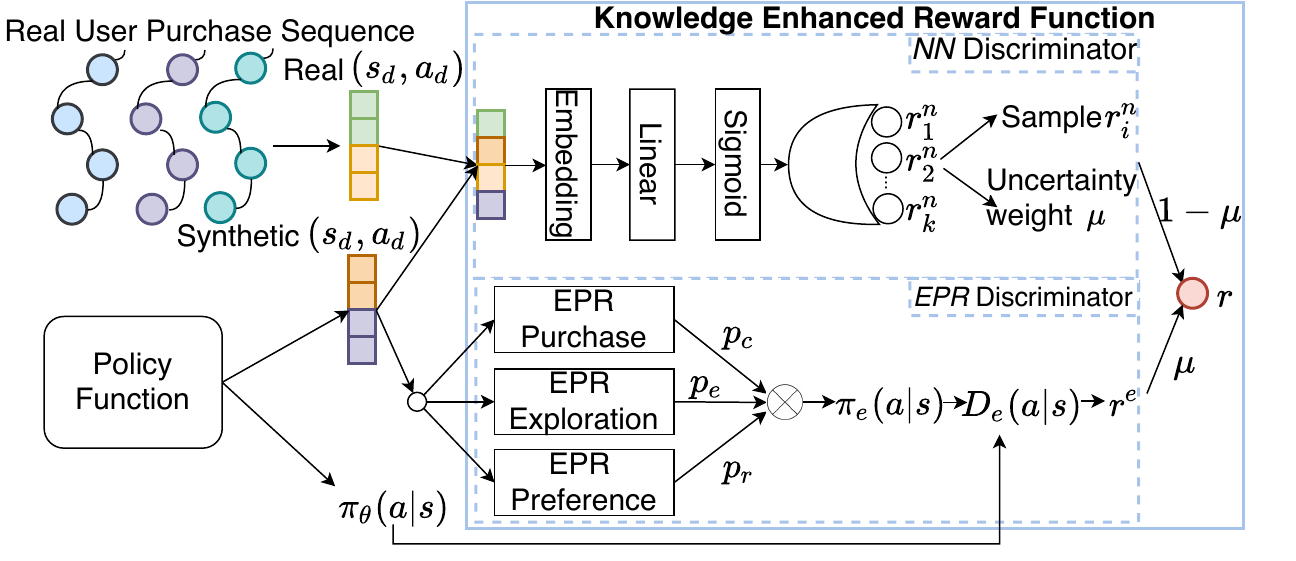}
\end{center}
\caption{Knowledge Enhanced Reward Function.} \label{fig:reward function}
\vspace{-3pt}
\end{figure}

\subsection{Knowledge Enhanced Reward Function}

In this section, we first introduce how the data-driven NN discriminator works. Then, we introduce the composition and principle of EPR discriminator based on EPR knowledge. Further, we propose an uncertainty-aware approach to synergize the two discriminators.

As shown in Figure~\ref{fig:reward function},
the input $(s_{d},a_{d})$ for the NN discriminator are sampled from both real-world and generated user consumption sequences $T^{r}$ and $T^{g}$, respectively.
Specifically, $s_{i,d}$ contains the consumption interval, visited number, the identifier of chosen store  and the feature of chosen store at $(i-1)$-th time slot, which can be denoted by $s_{i,d}=(l_{i-1},n_{i-1},id_{i-1}$, $c_{i-1},g_{i-1},v_{i-1},d_{i-1})$.
Further, $a_{i,d}$ contains the consumption interval, visited number and the feature of chosen store at $i$-th time slot, which can be denoted as $a_{i,d}=(l_{i},n_{i},id_{i},c_{i},g_{i},v_{i},d_{i})$. 
We denote the NN discriminator as $D_{\phi_{n}}$, which is parameterized by $\phi_{n}$. Then, the reward function $r^{n}$ of NN discriminator is $r^n=-log(1-D_{\phi_{n}})$ and $D_{\phi_{n}}$ is optimized based on the following loss function
\begin{equation}\label{equ:lossd}
   \mathcal{L}_D(\phi_n)=-\mathbb{E}_{\pi_{T^r}}[{\rm log}{D_{\phi_n}}(s_{d},a_{d})]-\mathbb{E}_{\pi_\theta}[{\rm log}(1-{D_{\phi_n}}(s_{d},a_{d}))],
\end{equation}
where $\mathbb{E}_{\pi_\theta}$ represents the expectation with respect to the user consumption samples by hierarchical policy function $\pi_\theta$.
In addition, $\mathbb{E}_{\pi_{T^r}}$ represents the expectation with respect to the state-action pairs obtained from the real-world user consumption samples.

However, due to the unbalanced interests of users to different stores,
there are fewer access samples to unpopular stores.
Therefore, the real-world user consumption sequence $T^{r}$ cannot cover all kinds of consumption behaviors.
As a result, it is challenging for the data-driven NN discriminator to judge and give right reward to transition $(s, a)$ for some sparse consumption behaviors, such as visiting unpopular stores.
Based on this intuition, we introduce a knowledge-driven approach to replicate the data-driven NN discriminator.
The EPR model is a classic probabilistic model~\cite{jiang2016timegeo} with high effectiveness in modeling user behaviors, such as generating behaviors, exploring new behaviors, and performing previous behaviors. Inspired by this, we propose an EPR discriminator based on EPR model to compensate for the deviations in the NN discriminator's reward estimation for the above mentioned $(s_{d},a_{d})$. 
As shown in Figure~\ref{fig:reward function}, the EPR discriminator is based on three probability models, including EPR Purchase, EPR Exploration and EPR Preference.

\para{EPR Purchase.}
It describes the probability distribution of the agent choosing to purchase in a store:
\begin{equation}\label{equ:teacherdisc}
   p_c(a_{i,d}^{c}=1|s_{i,d})=\alpha l_{i}^{-\beta},
\end{equation}
where $\alpha$ and $\beta$ are the distribution parameters that need to be fitted.

\para{EPR Exploration.}
It describes the probability distribution of the agent exploring a store he/she has never purchased before, 
\begin{equation}\label{equ:teacherdisc}
p_e(a_{i,d}^{e}=1|s_{i,d},a_{i,d}^{c}=1)=\rho n_{i}^{-\gamma},
\end{equation}
where $\rho$ and $\gamma$ are the distribution parameters that need to be fitted.

\para{EPR Preference.}
It follows two rules. Firstly, if the agent decides to return to previously purchased stores, he/she would choose a specific store $id_{i,d}=q$ with the following probability distribution: 
\begin{equation}\label{equ:teacherdisc}
p_r(a_{i,d}^p=q|s_{i,d},a_{i,d}^{c}=1,a_{i,d}^{e}=0)=v_{i,d}(q)/sum(v_{i,d}),
\end{equation}
where $v_{i,d}(q)$ is the number of historical purchase times at store $q$,
and $sum(v_{i,d})$ is the number of total purchase times by the agent.

Secondly, if the agent decides to explore to stores he/she has never been to,
he/she would choose a specific store $id_{i,d}=q$ with the following probability distribution:
\begin{equation}\label{equ:teacherdisc}
p_r(a_{i,d}^p=q|s_{i,d},a_{i,d}^{c}=1,a_{i,d}^{e}=1)=d_{i,d}(q)^{-\lambda},
\end{equation}
where $d_{i,d}(q)$ is the distance vector between the agent and the store $q$, and $\lambda$ is the distribution parameter that need to be fitted.

We denote the probability that the EPR framework outputs action $a$ given the state $s$ as $\pi_{e}(a|s)$ and the EPR discriminator as $D_{e}$.
Base on the above notations and multiplication theorem of probability, we can obtain that for a transition $(s_{i,d},a_{i,d})$,
we have $\pi_{e}(a_{i,d}|s_{i,d})=p_cp_ep_r$.

Inspired by the model design of GAIL and GAN ~\cite{ho2016generative,turner2019metropolis},
we design an EPR discriminator to classify between the conditional probability distribution $\pi_{\theta}(a|s)$
and $\pi_{e}(a|s)$ by 
\begin{equation}\label{equ:teacherdisc}
D_{e}(a|s)=\frac{\pi_{e}(a|s)}{\pi_{e}(a|s)+\pi_{\theta}(a|s)}.
\end{equation}
Therefore, the corresponding reward of EPR discriminator is $r^{e}(s,a)=-log(1-D_e(a|s))$.
We hope that the EPR discriminator can assist the NN discriminator to model the sparse purchase behavior $(s, a)$ in $T^{r}$, therefore it is important to know which $(s, a)$ is sparse and hard to model in $T^{r}$.
Inspired by the existing works on model uncertainty~\cite{da2020uncertainty,oh2019sequential},
we propose a multi-head uncertainty-aware method to combine the NN discriminator with the EPR discriminator, as shown in Figure~\ref{fig:reward function}.
The key idea is to utilize the uncertainty value of the reward output by the discriminator to determine whether the input transition $(s,a)$ is sparse in $T^{r}$. 

We first refer to the bootstrapped DQN~\cite{osband2016deep} and design a multi-head reward output  for the NN discriminator.
As shown in Figure~\ref{fig:reward function}, the NN discriminator outputs a k-head reward: $r_{1}^{n},r_{2}^{n}...r_{k}^{n}$. For the process of generating a user purchase sequence for $\pi_{\theta}$, we randomly sample one head from $k$ heads to provide the reward for $\pi_{\theta}$. Then, based on the uncertainty value proposed in~\cite{da2020uncertainty},
we set an uncertainty weight $u$ to weight the rewards of the NN discriminator and the EPR discriminator by
\begin{equation}\label{equ:uncertainty}
u=\mathrm{sigmoid}(\mathrm{var}(r^{n})),
\end{equation}
where $\mathrm{var(\cdot)}$ is variance of the vector.

When the transition $(s,a)$ is sparse in $T^{r}$ and the uncertainty is higher, the output reward should be more dependent on the EPR discriminator. Therefore, the final reward $r(s,a)$ of knowledge enhanced reward function can be computed as follows.
\begin{equation}\label{equ:r}
r(s,a)=(1-u)r_{i}^{n}(s,a)+ur^{e}(s,a).
\end{equation}
We summarize details of the training process of  EPR-GAIL in Algorithm 1 of Appendix \ref{sec:algorithm}.



\section{Experiments}

In this section, we conduct extensive simulation experiments based on two real-world user consumption datasets to verify the effectiveness of EPR-GAIL framework.


\subsection{Datasets}
Based on a major online consumer platform in China,
we collect two datasets of real-world user consumption sequences across two capital cities (i.e., Beijing and Guiyang), respectively. The details of the two dataset are introduced in Appendix \ref{sec:Datasets}.

\subsection{Experimental Settings}
\para{Baseline Algorithms.} To verify the effectiveness of EPR-GAIL, we compare its performance with 6 state-of-the-art baseline methods: IO-HMM~\cite{yin2017generative}, EPR~\cite{jiang2016timegeo}, SeqGAN~\cite{yu2017seqgan}, MoveSim~\cite{FengYXYWL20}, BC~\cite{torabi2018behavioral}, GAIL~\cite{ho2016generative}.  The details of baselines are introduced in Appendix \ref{sec:bl}.

\para{Statistical Evaluation Metrics.}
To evaluate the quality of the generated consumption sequences,
we measure the similarity between generated sequences and real-world
sequences by adopting the following indicators as metrics. \textbf{Consumption}: This indicator measures the behavior similarity between generated sequences and real-world sequences on whether consumption behaviors  actually occur or not; \textbf{Exploration}: This indicator measures the behavior similarity between the generated sequences and real-world sequences on exploring new stores; \textbf{Price}: This metric measures the similarity of the average price of selected stores between the generated sequences and real-world sequences; \textbf{Distance}: This metric measures the similarity of the distance of the selected stores between the generated sequences and real-world sequences; \textbf{Type}: This metric measures the similarity of selected stores in generated sequences and real-world sequences; \textbf{Identifier}: This metric measures the similarity identifiers of selected stores in  generated sequences and real-world sequences.

Note that all metrics are expressed by probability distributions with each consumption sequence.
To intuitively measure the similarity between the generated sequences and the real-world sequences, we further adopt the Jensen-Shannon Divergence (JSD)~\cite{thomas2006elements} as a key metric in computation.
Specifically, for two distribution $\boldsymbol{p}$ and $\boldsymbol{q}$, the JSD between them can be computed by
\begin{equation}\label{equ:JSD}
{\rm JSD}(\boldsymbol{p},\boldsymbol{q})=\frac{1}{2}{\rm KL}(\boldsymbol{p}||\frac{\boldsymbol{p}+\boldsymbol{q}}{2})+\frac{1}{2}{\rm KL}(\boldsymbol{q}||\frac{\boldsymbol{p}+\boldsymbol{q}}{2}),
\end{equation}
where ${\rm KL}(\cdot||\cdot)$ is the Kullback-Leibler divergence~\cite{thomas2006elements}. 

\subsection{Performance Comparison}

\begin{table*}[t]
    \centering
    \vspace{-3pt}
    \caption{Performance comparison over all baselines in two datasets.}
    \resizebox{1.0\textwidth}{!}
    {
    \begin{tabular}{c||cccccc|cccccc}
         \Xhline{1pt} \textbf{Dataset}& \multicolumn{6}{c|}{\textbf{Beijing}}
        & \multicolumn{6}{c}{\textbf{Guiyang}}
        \\
        Metrics (JSD) &Consumption &Exploration &Price & Distance &Type &Identifier &Consumption &Exploration &Price & Distance&Type&Identifier  \\
        \hline \hline 
        IO-HMM&0.47&0.41&0.39&0.38&0.17&0.33&0.57&0.51&0.39&0.41&0.35&0.37\\
        EPR&0.13&0.12&0.48&0.21&0.38&0.27&0.14&0.15&0.47&0.26&0.34&0.35\\
        SeqGAN&0.39&0.28&0.31&0.32&0.29&0.35&0.41&0.38&0.36&0.45&0.33&0.37\\
        MoveSim&0.15&0.17&0.35&0.31&0.11&0.22&0.17&0.18&0.37&0.33&0.14&0.25\\
        \hline 
        BC&0.58&0.57&0.47&0.41&0.15&0.31&0.55&0.59&0.42&0.39&0.38&0.33\\
        GAIL&0.34&0.25&0.27&0.36&0.19&0.25&0.37&0.35&0.39&0.41&0.29&0.30\\
        \hline
        EPR-GAIL&\textbf{0.0004}&\textbf{0.0018}&\textbf{0.21}&\textbf{0.17}&\textbf{0.02}&\textbf{0.013}&\textbf{0.0008}&\textbf{0.0029}&\textbf{0.23}&\textbf{0.16}&\textbf{0.03}&\textbf{0.033}\\
        \hline
        Percentage&\textbf{96.92\%}&\textbf{98.50\%}&\textbf{22.22\%}&\textbf{19.04\%}&\textbf{81.81\%}&\textbf{94.09\%}&\textbf{99.42\%}&\textbf{98.07\%}&\textbf{36.11\%}&\textbf{38.46\%}&\textbf{78.57\%}&\textbf{86.80\%}\\
        \Xhline{1pt}
    \end{tabular}
    }
    \label{tab:all_result}
    \vspace{-2mm}
\end{table*} 

\begin{table*}[t]
    \centering
    \caption{Experimental results in the ablation study.}
    \resizebox{1\textwidth}{!}
    {
    \begin{tabular}{c||cccccc|cccccc}
        \Xhline{1pt} \textbf{Dataset}
        & \multicolumn{6}{c|}{\textbf{Beijing}}& \multicolumn{6}{c}{\textbf{Guiyang}} \\
        Metrics (JSD) &Consumption &Exploration &Price & Distance &Type &Identifier &Consumption &Exploration &Price & Distance &Type &Identifier \\
        \hline \hline
        No-pretrain&0.21&0.19&0.29&0.27&0.12&0.13&0.19&0.23&0.27&0.23&0.16&0.13\\
        No-self-attention&0.013&0.027&0.25&0.19&0.05&0.113&0.018&0.039&0.33&0.20&0.09&0.11\\
        No-LSTM&0.011&0.018&0.28&0.20&0.09&0.093&0.027&0.054&0.27&0.19&0.09&0.101\\
        \hline 
        EPR-GAIL &\textbf{0.0004}&\textbf{0.0018}&\textbf{0.21}&\textbf{0.17}&\textbf{0.02}&\textbf{0.013}&\textbf{0.0008}&\textbf{0.0029}&\textbf{0.23}&\textbf{0.16}&\textbf{0.03}&\textbf{0.033}\\
        \Xhline{1pt}
    \end{tabular}
    }
    \label{tab:ablation}
\end{table*}

\subsubsection{\textbf{Overall Performance}}
The overall performance comparison of EPR-GAIL and all baseline methods on generating synthetic user consumption sequences are shown in Table~\ref{tab:all_result}.
Based on the evaluation results, we make the following observations. First, among all conventional sequence generation methods,
MoveSim achieves the best results,
as it generates user consumption sequences based on contextual information between users and stores.
Besides, the EPR method achieves the best performance on JSD Consume and JSD Explore,
as it focuses more on modeling users' consumption and exploration behaviors.
Second, for imitation learning-based methods,
the GAIL model performs better than BC model.
The key reason is that user consumption sequence generation is a dynamic process,
meanwhile, the BC model will produce large errors when new states emerge. Third, in comparison with all baseline methods, the proposed EPR-GAIL stably shows the best performance under all metrics, achieving at least 19.04\% performance improvement.
Correspondingly, EPR-GAIL exhibits strong capability to model user consumption behavior sequences with high precision and fidelity. To conclude, compared with the state-of-the-art baselines,
EPR-GAIL achieves notable performance gains in user consumption sequence generation.
More importantly, the proposed EPR-GAIL shows its superiority of modeling complex user consumption behaviors.

\subsubsection{\textbf{Ablation Study}}
To provide a comprehensive understanding of each key component's role in EPR-GAIL framework,
we conduct an ablation study to investigate the performance of EPR-GAIL's variants.
In general, we develop three different variants of the EPR-GAIL model as follows.
\emph{No-pretrain} is a variant of EPR-GAIL without the pre-training process.
\emph{No-self-attention} is a variant of EPR-GAIL without the self-attention module.
\emph{No-LSTM} is a variant of EPR-GAIL without the LSTM module.
The experimental results for the ablation study are presented in Table~\ref{tab:ablation} and our key findings are as follows.
First, the pre-training process is particularly important to guarantee the performance of EPR-GAIL, as No-pretrain variant shows significant performance degradation across all JSD metrics.
Second, the self attention module and the LSTM module have almost equal contributions to the performance gain of EPR-GAIL, as No-self-attention variant and \emph{No-LSTM} variant shows similar JSD values overall different metrics.
Third, without the pre-training process, the Purchase agent and the Exploration agent will not be able to provide valid information to the Preference agent,
which will lead to EPR-GAIL to generate low-quality consumption sequences (as revealed from larger JSD metrics.)

\subsubsection{\textbf{Practical Demonstrations}}
To verify the effectiveness of consumption sequences generated by EPR-GAIL,
we conduct experiments with two real-world applications.

\begin{figure}[t]
\centering
\subfigure[Beijing]{\includegraphics[width=.336\textwidth]{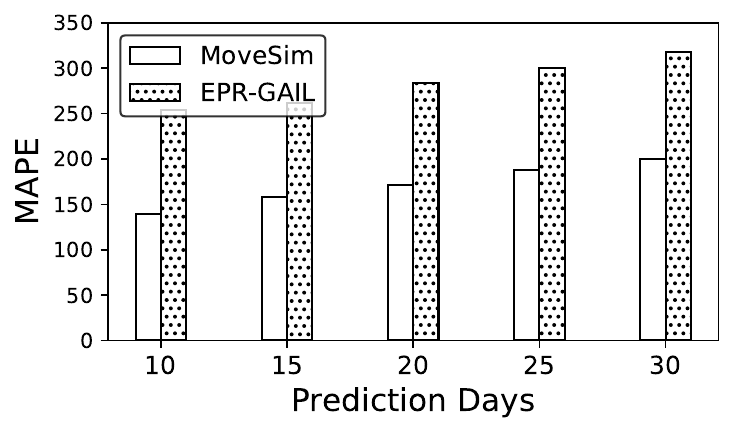}}
\subfigure[Guiyang]{\includegraphics[width=.336\textwidth]{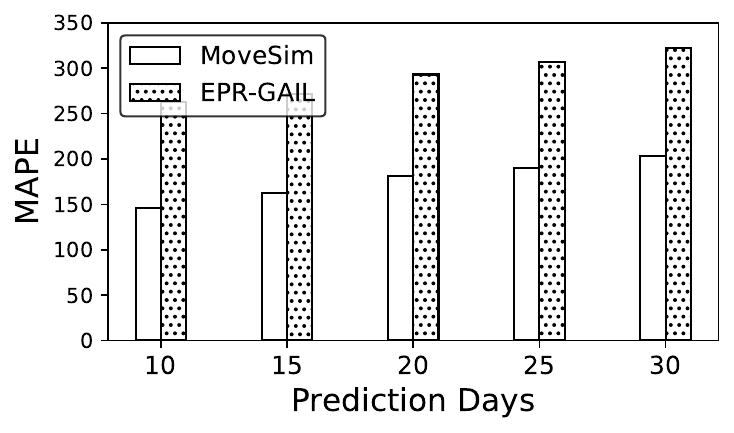}}
\caption{Store sales prediction on different datasets.} 
\label{fig:sale}
\vspace{-3pt}
\end{figure}

\para{Sales Prediction.}
Sales prediction is widely useful for a variety of applications,
as it can assist investment selection and business analysis of online stores.
However,as the number of online stores will change when the new stores go online,
traditional forecasting methods cannot adapt to such changes in the decision space.
Therefore, we propose a simulation-based method to predict store sales.
On the one hand, we generate each user's future consumption sequences based on EPR-GAIL and the baseline MoveSim.
On the other hand, we further aggregate consumption of all users from the store's perspective to derive the ground-truth sales of each store.
We compare the performance of EPR-GAIL with the best baseline MoveSim on the Mean Average Percentage Error (MAPE) metric by varying the number of days, and the results are shown in Figure~\ref{fig:sale}.
It is worth finding that the sales prediction results are more accurate
(\emph{i.e.}, at least 35.29\% improvement) based on the generated consumption sequences from EPR-GAIL than the results assisted by MoveSim.

\begin{figure}[t]
\centering
 \subfigure[Beijing]{\includegraphics[width=.336\textwidth]{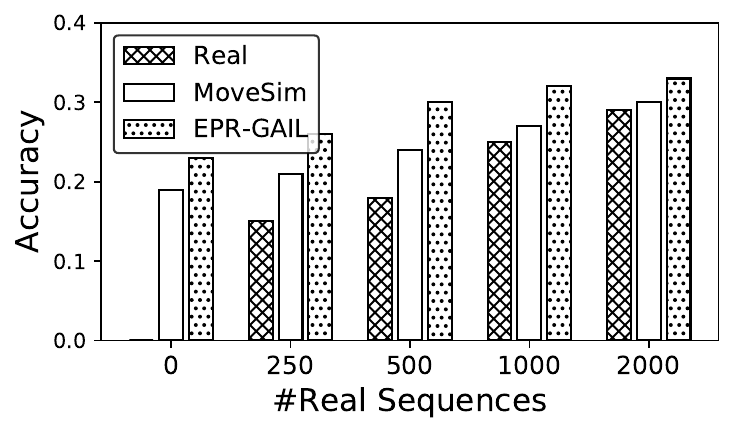}}
\subfigure[Guiyang]{\includegraphics[width=.336\textwidth]{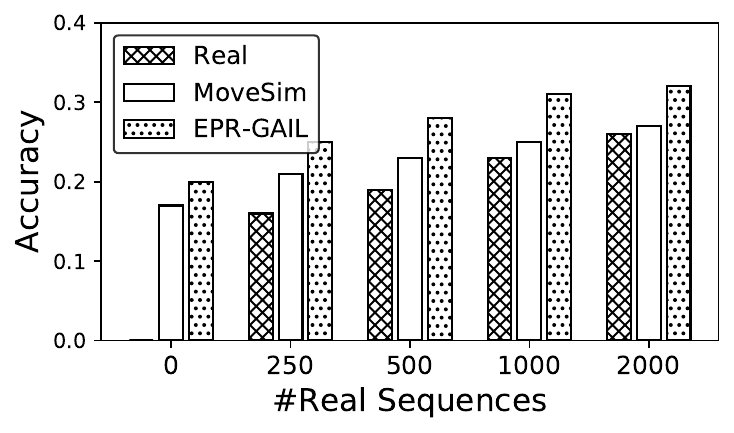}}
\vspace{-3pt}
\caption{Location recommendation on different datasets.} \label{fig:reco}
\end{figure}

\para{Location Recommendation.} 
Location recommendation is a commonly used application by online platforms,
and it only tends to succeed with large amounts of data.
Yet, concerning the privacy and security issues, the real-world user consumption behavior data is not readily available to any third-party counterparts at full scale.
Therefore, we explore to train the location recommendation model by combining
the synthetic user consumption behavior data with a small amount of real-world data.
Specifically, we exploit the classic collaborative filtering algorithm as the base model,
and further combine different amounts of real-world data with artificially generated data to train the recommendation model.
Overall, we compare EPR-GAIL with the best baseline MoveSim in generating 2,000 synthetic sequences.
Figure~\ref{fig:reco} illustrates the accuracy of location recommendation results in two different cities using the above combined data.
We can find that EPR-GAIL shows robust performance improvement over MoveSim with at least 11.19\% improvement, despite of different proportions of synthetic sequences to real-world sequences.

In addition, we conducted experimental explorations on three aspects: the fitting of the EPR model to real user consumption behavior (Appendix \ref{sec:Knowledge Discovery}), the impact of EPR knowledge on downstream tasks (Appendix \ref{sec:Performance of EPR}), and the impact of replacing EPR knowledge with other expert knowledge on the performance of downstream tasks (Appendix \ref{sec:Alternatives}) to further verify the effectiveness and scientificity of EPR knowledge.

\section{Related Work}
The earliest methods of human behavior generation are formulaic methods based on statistical laws~\cite{stouffer1940intervening,simini2012universal}.
These methods are often applicable in many occasions and have strong robustness but are hard to model the latent properties within the human behavior data. 
Deep learning-based methods have been applied to human behavior generation in recent years, including variational auto-encoder (VAE)~\cite{huang2019variational},
generative adversarial network (GAN)~\cite{feng2018deepmove,ouyang2018non,kulkarni2018generative,Liu2018trajGANsU}.
Nevertheless,
these methods are still difficult to model complex human behaviors.
With the rise of imitation learning, there has been some generative work that models human behavior as human behavioral decisions~\cite{pan2020xgail,zhang2019unveiling,wu2020joint, ZangTrajGAIL, SeongjinChoi, ISrein}.
However, these works are particularly dependent on collected expert data, which makes it difficult to use in practical applications.

\section{Conclusion}
In this study, we have proposed EPR-GAIL, a novel framework that integrates the generative adversarial imitation learning method with the exploration and preferential return model.
Our core idea is to model user consumption behaviors as a complex EPR decision process, with successive decisions of purchase, exploration and preference.
We have built the EPR-GAIL framework with three functioning modules,
including a decision-making feature extraction module, a hierarchical policy function module, and a knowledge-enhanced reward function module. In addition, EPR-GAIL still has the risk of some user privacy leaks. Therefore, we will take it into account in our model in future research.

We have further collected two real-world datasets, covering online consumption behaviors of more than ten thousand users over three months, respectively.
The experimental results with the above two datasets have demonstrated that EPR-GAIL framework significantly outperforms state-of-the-art methods in user consumption behavior simulation by over 19\% in terms of data fidelity.
The generated consumption behavior data by the EPR-GAIL has further enhanced the performance of sale prediction and location recommendation by up to 35.29\% and 11.19\%, respectively.

\newpage
\bibliographystyle{ACM-Reference-Format}
\bibliography{bibliography}

\newpage
\appendix
\section{Appendix}

\subsection{Decision-Making Feature Extraction}\label{sec:Feature Extraction}
\begin{figure} [ht]
\begin{center}
\includegraphics*[width=0.69\textwidth]{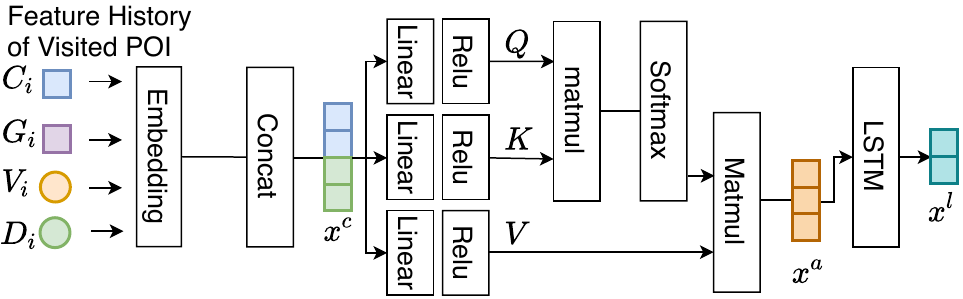}
\end{center}
\caption{Decision-making feature extraction module.} \label{fig:feature}
\end{figure}
To capture complex sequential transition regularities in user consumption sequences,
we design a decision-making feature extraction module with both a self-attention layer~\cite{vaswani2017attention} and an LSTM layer.
As shown in Figure~\ref{fig:feature}, the feature extraction module first embeds the feature history of stores that each user has consumed before the $i$-th time interval, which contains the category information $C_{i}$, the average price information $G_{i}$, the visit information $V_{i}$ and the historical user-store distance $D_{i}$.
It then concatenates the above embeddings into a dense representation vector $x^{c}$,
which can be derived by

\begin{equation}\label{equ:pate}
x_{i}^{c}=Relu([C_{i}W_{c},G_{i}W_{g},V_{i}W_{v},D_{i}W_{d}]),
\end{equation}
where $x_{i}^{c}$ is the vector representation of $i$-th time interval,
and $W_{c},W_{g},W_{v},W_{d}$ are learnable parameters of embedding table.
Based on the vector representation $x^{c}$,
we further design a self-attention layer to capture the relationship between different features of consumed stores in the same time slot, so as to model each user's consumption behavior.
Based on the module structure in Figure~\ref{fig:feature},
$x_{i}^{c}$ is first projected with three non-linear operations into three vectors,
including the query vector $Q_{i}$, the key vector $K_{i}$, and the value vector $V_{i}$.
Then, a scaled dot-product attention is applied on them to obtain a weighted sum of value vectors as the overall feature representation of $i$-th time interval.
The above vectors can be computed by

\begin{equation}\label{equ:pate}
Q=Relu(x^{c}W_{Q}), K=Relu(x^{c}W_{K}), V=Relu(x^{c}W_{V}),
\end{equation}

\begin{equation}\label{equ:pate}
x^{a}=\mathrm{softmax}(QK/\sqrt{d})V,
\end{equation}
where $x^{a}$ is the output of the self-attention layer and
$W_{Q}$, $W_{K}$, $W_{V}$ are learnable parameters.
Furthermore, we apply an LSTM layer to project feature representation sequences embedded by the self-attention layer,

\begin{equation}\label{equ:pate}
x^{l}=LSTM(x^{a}),
\end{equation}
where $x^{l}$ is the output of LSTM and it serves as the feature representation for
the decision-making process of the hierarchical policy function.

\subsection{The pseudo algorithm of EPR-GAIL}\label{sec:algorithm} 

We summarize details of the training process of EPR-GAIL in Algorithm~\ref{alg:framework}. From the algorithm, we can first observe that a batch of synthetic sequences and real-world sequences are sampled to train the NN discriminator (lines 8-9). The synthetic sequences are regarded as negative samples and real-world sequences are regarded as positive samples to train NN discriminator via an Adam Optimizer~\cite{kingma2014adam} (line 10). Then, we calculate a batch of rewards for the state-action pairs in the sampled synthetic sequences through the sampled reward head of NN discriminator and calculated uncertainty weight (lines 4-5,7).
Finally, we train the hierarchical policy function by maximizing the expectation of reward via a reinforcement learning algorithm called Proximal Policy Optimization (PPO)~\cite{schulman2017proximal} (line 12).

\renewcommand{\algorithmicrequire}{\textbf{Input:}}
\begin{algorithm}[htbp]
\caption{EPR-GAIL}\label{alg:framework}
\begin{algorithmic}[1]
\REQUIRE
Real-world user purchase sequence $T^{r}$.
\STATE
Initialize $\pi_{\theta}$, $D_{\phi_n}$ with random weights;
\FOR{$i$=0,1,2...}
\STATE 
Generate a batch of sequences $T^{g}_{i}$ $\sim$$\pi_\theta$;
\STATE 
Random sample a reward head $r_{i}^n$ of neural network-based discriminator;
\STATE
Calculate the uncertainty weight $u$ based on (\ref{equ:uncertainty});
\FOR{$k=0,1,2...$}
\STATE
Calculate a batch of rewards $r(s,a)$ based on (\ref{equ:r}) for each state-action pair in sequences $T^g_i$;
\STATE
Sample synthetic sequences $T^g_{ki}$ based on $\pi_\theta$;
\STATE
Sample real-world sequences $T^r_k$ from $T^r$  with same batch size;
\STATE
Update $D_{\phi_n}$ based on (\ref{equ:lossd}) Adam Optimizer with the positive samples $T^r_k$ and negative samples $T^g_{ki}$.
\ENDFOR
\STATE
Update $\pi_\theta$ by maximizing the expectation of reward $\mathbb{E}_{\pi_\theta}(r(s,a))$ via the PPO method;
\ENDFOR
\end{algorithmic}
\end{algorithm}

\subsection{Datasets}\label{sec:Datasets} 
We summarize the metadata of the above two datasets in Table~\ref{table:data1}. The detail of each dataset is as follows.
\begin{itemize}
\item \textbf{Beijing Dataset}. This dataset covers user consumption behavior sequences of 10,152 users at Beijing, during May 2021 to July 2021. In total, it covers 6,903 stores operating on the online consumer platform.
\item \textbf{Guiyang Dataset}. This dataset covers user consumption behavior sequences of 10,060 users at Beijing, during April 2021 to June 2021. In total, it covers 7,931 stores operating on the online consumer platform.
\end{itemize}

Each dataset covers online user activities for two months and contains a variety of information,
including the online consumption sequences of platform users, the latitude and longitude information of users,
the category of stores, the average consumption price of stores, as well as the latitude and longitude information of stores. The metadata of the above two datasets is summarized in Table~\ref{table:data1}.

\begin{table}[t]
\centering
\caption{Statistic of Datasets.}\label{table:data1}
\vspace{-2mm}
\resizebox{2.5in}{!}{
\begin{tabular}{ccc}
\hline
\textbf{Datasets} & \textbf{\emph{Beijing}} & \textbf{\emph{Guiyang}} \\\hline
Location & Beijing & Guiyang \\
Time Span & \begin{tabular}[c]{@{}c@{}}5/1/2021-1/7/2021\end{tabular} & \begin{tabular}[c]{@{}c@{}}4/1/2021-6/1/2021\end{tabular} \\
Number of Users & 10,152 & 10,060 \\
Number of Stores & 6,903 & 7,931 \\\hline
\end{tabular}}
\end{table}

\subsection{Baseline Methods}\label{sec:bl} 
We provide the details of baseline methods in our experiments as follows.
We first compare our method with four baselines for sequence generation,
\begin{itemize}
\item \textbf{{IO-HMM}~\cite{yin2017generative}}: This model is based on the  Hidden Markov model, which generates synthetic sequences with external context information of both users and stores.

\item \textbf{{EPR}}~\cite{jiang2016timegeo}: This model is a rule-based probabilistic model, which models the users' consumption behaviors such as purchase, exploring a new store or returning to a store which has been consumed before.

\item \textbf{{SeqGAN}}~\cite{yu2017seqgan}: Different from standard GANs~\cite{goodfellow2014generative}, this generator is designed to generate sequences through training with RL algorithm.

\item \textbf{{MoveSim}}~\cite{FengYXYWL20}: This model is a state-of-the-art trajectory generation algorithm based on GAN. To adapt to our research problem, we change the action selection of MoveSim model to the selection of the store, and further change the distance constraint of mobility to the distance constraint between users and stores.

We then compare our method with two imitation learning  baselines,

\item \textbf{{BC}}~\cite{torabi2018behavioral}: This model is a supervised imitation learning method, which takes the current state as an input and the choice of the next moment as a label.

\item \textbf{{GAIL}}~\cite{ho2016generative}: This model generates user consumption sequences by modeling the sequence generation as a human decision-making process. In general, a GAIL model includes a discriminator module and a policy module. 
\end{itemize}

\begin{figure}[htbp]
\centering
\subfigure[Fitting of EPR Consume with $R^2=0.98$.]{
    \label{pzdim}
    \includegraphics[width=0.45\textwidth]{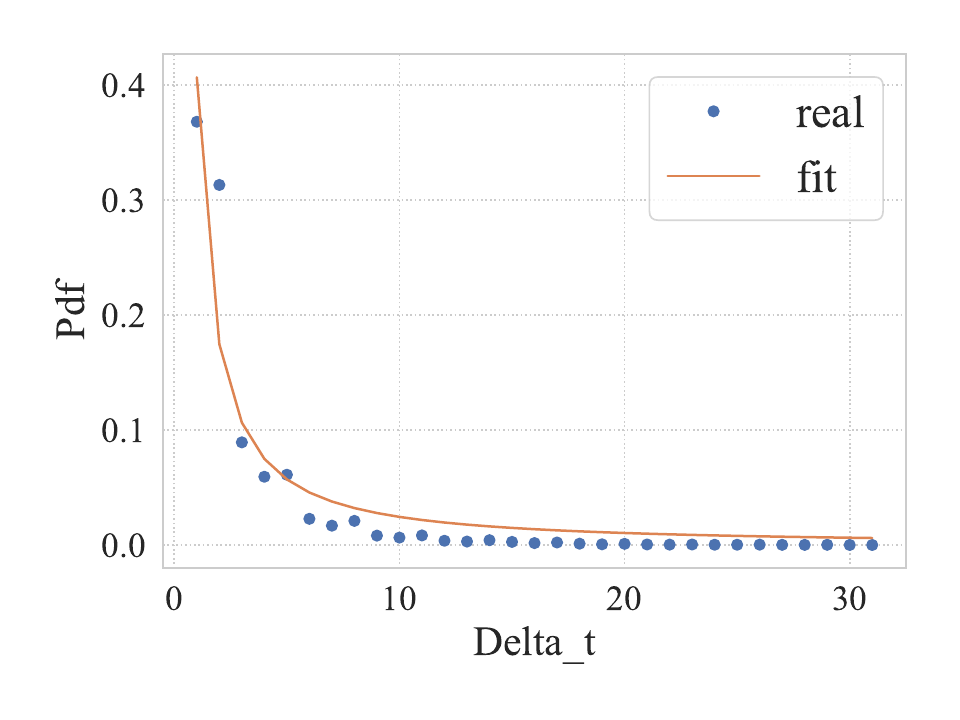}
}
\subfigure[Fitting of EPR Explore with $R^2=0.66$.]{
    \label{beta}
    \includegraphics[width=0.45\textwidth]{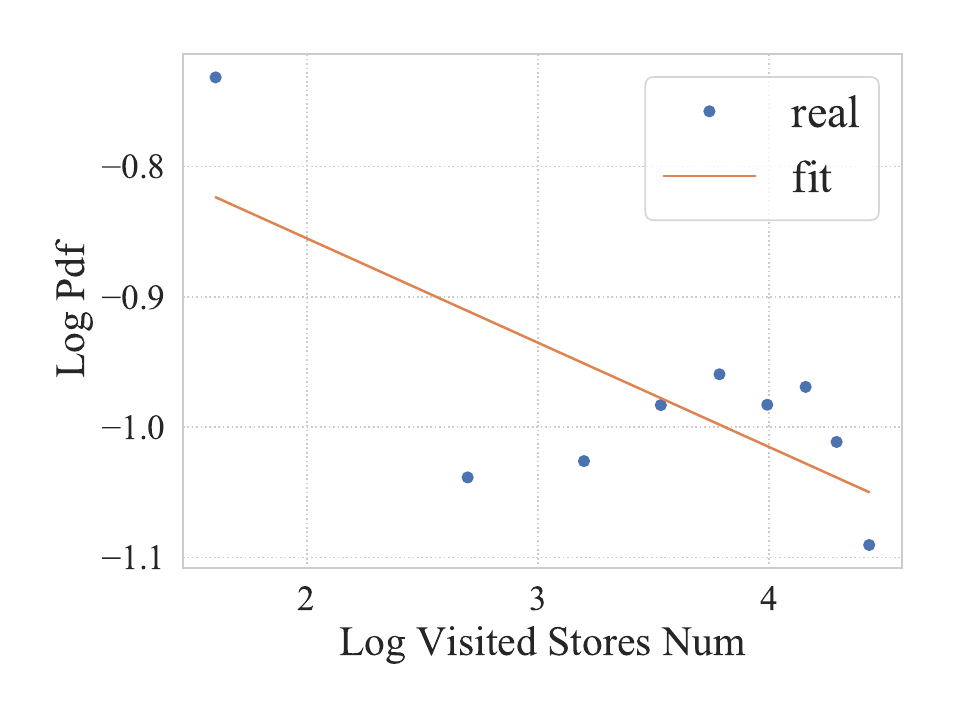}
}

\subfigure[Fitting of EPR Preference when a user choose to explore a store with $R^2=0.72$.]{
    \label{layers}
    \includegraphics[width=0.45\textwidth]{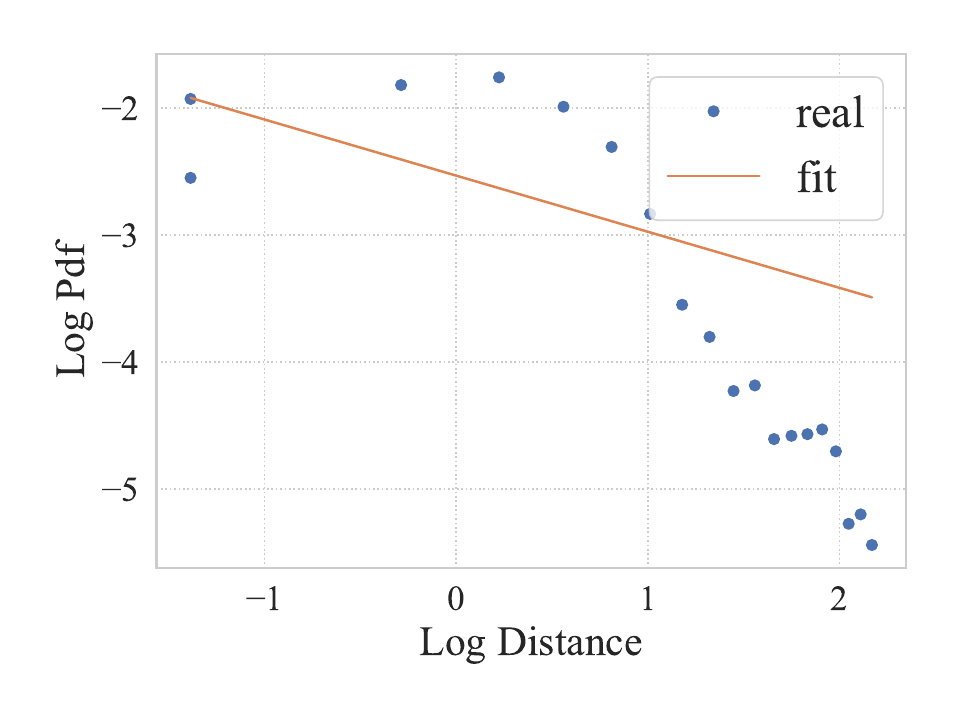}
}
\subfigure[Fitting of EPR Preference when a user choose to return a store he/she has consumed before with $R^2=0.99$.]{
    \label{filters}
    \includegraphics[width=0.45\textwidth]{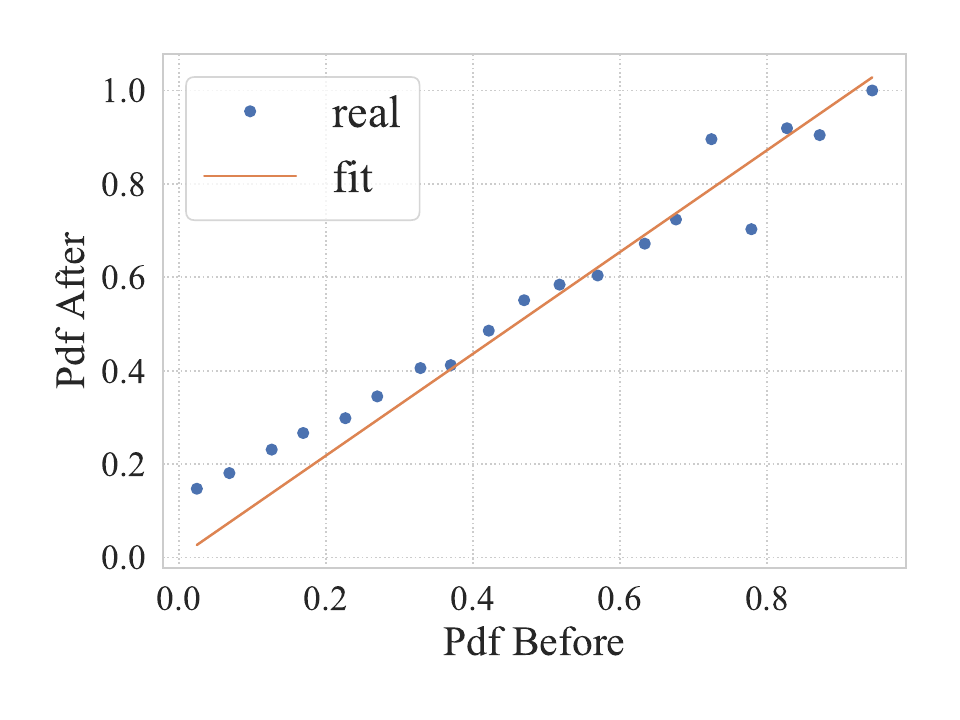}
}
\caption{Discovery of EPR knowledge in Beijing dataset.}
\label{EPR know}
\end{figure}

\begin{figure}[t]
  \centering
  \includegraphics[width=0.55\textwidth]{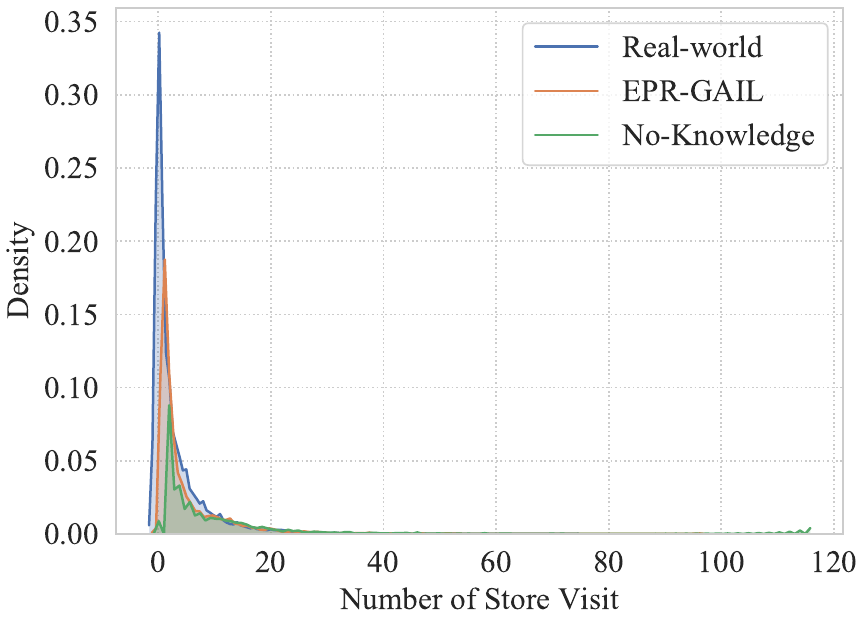}
  \caption{Distribution of customers' visits to stores}
  \label{Fig:standard deviation}
\end{figure}

\subsection{EPR Knowledge Discovery}\label{sec:Knowledge Discovery}
To verify that EPR knowledge can help GAIL generate sequences better, we verify the EPR model referring  to timegeo ~\cite{jiang2016timegeo} on the Beijing dataset, which is shown in Fig.~\ref{EPR know}.
From the figure and coefficient of determination for fitting with real data ($R^2$) we can observe that the fit of the real data roughly conforms to the trend of the EPR model, which has been discussed in Sec 3.4.

\subsection{Performance of EPR Knowledge}\label{sec:Performance of EPR}
\begin{figure}[t]
\centering
\subfigure[Beijing]{\includegraphics[width=.336\textwidth]{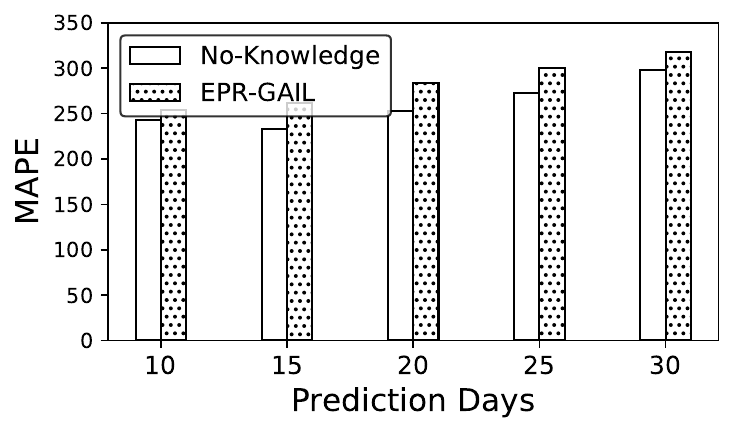}}
\subfigure[Guiyang]{\includegraphics[width=.336\textwidth]{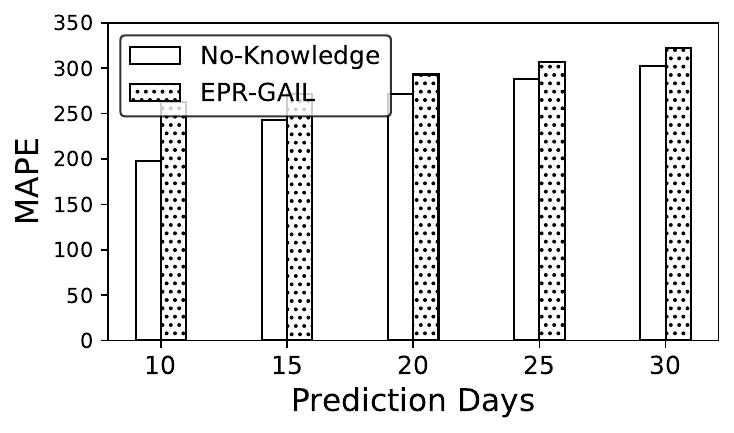}}
\caption{The impact of EPR knowledge on the sales prediction task.} \label{fig:task1}
\end{figure}

\begin{figure}[t]
\centering
\subfigure[Beijing]{\includegraphics[width=.336\textwidth]{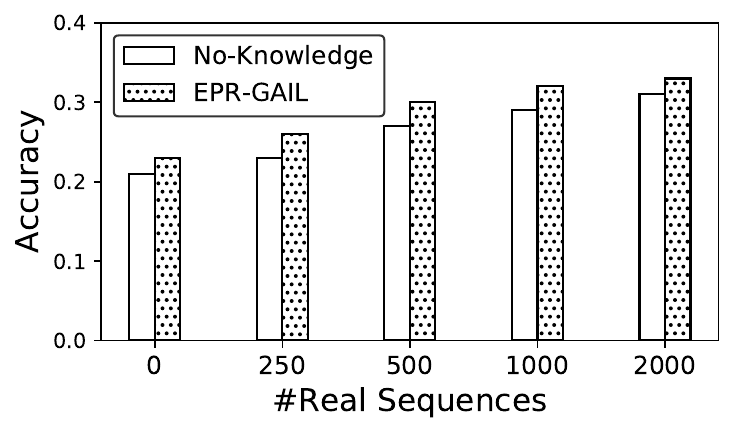}}
\subfigure[Guiyang]{\includegraphics[width=.336\textwidth]{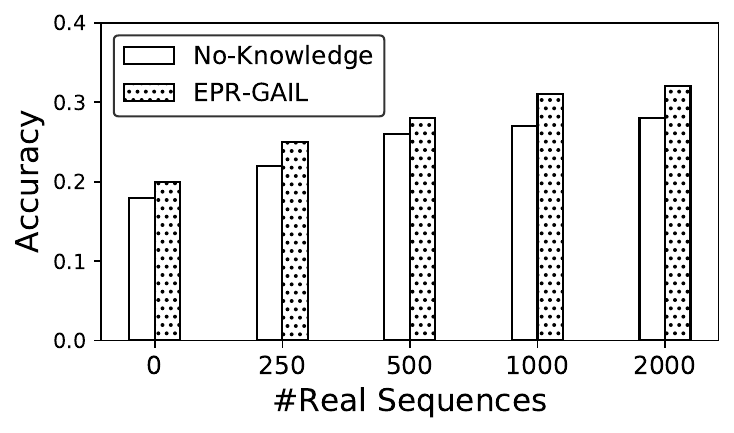}}
\caption{The impact of EPR knowledge on the location recommendation task.} \label{fig:task2}
\end{figure}

At last, to quantify the performance enhancement brought by the EPR knowledge to the GAIL model in user behavior sequence generation, we verify the EPR model introduced in Sec 3.4 by referring to the TimeGeo modeling framework~\cite{jiang2016timegeo} on the Beijing dataset. We fit the distribution of real data with the EPR probability model and show the fit based on the coefficient value $R^2$. Figure~\ref{EPR know} shows our result and it can be observed user consumption datas are roughly consistent with the EPR probability model. For example, we draw the distribution of the consumption time interval $delta_t$ in Figure ~\ref{EPR know} (a), which presents an obvious power-law distribution form and  is consistent with ERP Purchase described in equation (10).  We further verify the importance of EPR knowledge for GAIL in predicting the store visits caused by the potential purchases of new users. Specifically, we train EPR-GAIL on a dataset of 2,000 users,
and then use it to generate user consumption sequences on a dataset of 10,000 users.
By counting the probability distribution of visit frequency to different stores,
we compare the distribution of our generated data to the distribution of real-world data, and the distribution of generated data by a variant of EPR-GAIL without the EPR knowledge.
Based on results shown in Figure~\ref{Fig:standard deviation}, we can find that with EPR knowledge,
EPR-GAIL can better model user consumption behaviors that are not covered by the training set.
We also compare the performance of EPR-GAIL with its No-knowledge variant on sales prediction task and location recommendation task.
As shown in Figure~\ref{fig:task1} and Figure~\ref{fig:task2}, it can be found that the EPR-GAIL with EPR knowledge can achieve a great gain for the downstream applications,
thereby verifying the effectiveness of the EPR knowledge.

\begin{table*}[ht]
    \centering
    \caption{The alternatives of EPR knowledge on the sales prediction task.}
    \resizebox{1.\textwidth}{!}
    {
    \begin{tabular}{c||ccccc|ccccc}
         \Xhline{1pt} \textbf{MAPE}& \multicolumn{5}{c|}{\textbf{Beijing}}
        & \multicolumn{5}{c}{\textbf{Guiyang}}
        \\
        Prediction Days &10 &15 &20 &25 &30 &10 &15 &20 &25 & 30  \\
        \hline \hline 
        No-Knowledge&253.9&262.1&283.6&300.7&317.7	&262.8&271.4&292.1	&306.7	&322.4\\
        Distance&250.1&261.4&280.1&296.4&311.9&255.3&265.9&284.7&303.3&318.9\\
        \hline 
        Price&249.8&255.5&273.2&288.1	&308.4	&245.6&264.3&280.5&298.1	&314.5\\
        Popularity&248.3&252.1&266.9	&280.1&306.1&233.4&255.1&278.6&295.6&311.7\\
        \hline
        EPR-GAIL&\textbf{243.1}&\textbf{232.6}&\textbf{252.7	}&\textbf{272.3}&\textbf{298.4	}&\textbf{198.2}&\textbf{243.5}&\textbf{271.7}&\textbf{288.5}&\textbf{302.3
}\\
        \Xhline{1pt}
    \end{tabular}
    }
    \label{tab:alternatives_sales}
\end{table*}

\begin{table*}[h]
    \centering
    \caption{The alternatives of EPR knowledge on the location recommendation task.}
    \resizebox{1.0\textwidth}{!}
    {
    \begin{tabular}{c||ccccc|ccccc}
         \Xhline{1pt} \textbf{Accuracy}& \multicolumn{5}{c|}{\textbf{Beijing}}
        & \multicolumn{5}{c}{\textbf{Guiyang}}
                \\
        Real Sequences &0 &250 &500 &1000 &2000 &0 &250 &500 &1000 &2000  \\
        \hline \hline 
        No-Knowledge&0.21&0.23&0.27&0.29&0.31&0.18&0.22&0.26&0.27&0.28\\
        Distance&0.21&0.22&0.25&0.29&0.31&0.16&0.22&0.27&0.25&0.28\\
        \hline 
        Price&0.20&0.23&0.28&0.28&0.30&0.17&0.21&0.25&0.26&0.29\\
        Popularity&0.22&0.24&0.29&0.30&0.18&0.23&0.27&0.27&0.27&0.30\\
        \hline
        EPR-GAIL&\textbf{0.23}&\textbf{0.26	}&\textbf{0.30	}&\textbf{0.32	}&\textbf{0.33}&\textbf{0.20}&\textbf{0.25	}&\textbf{0.28}&\textbf{0.31}&\textbf{0.32
}\\
        \Xhline{1pt}
    \end{tabular}
    }
    \label{tab:alternatives_recommendation}
\end{table*}

\subsection{Alternatives of EPR Knowledge}\label{sec:Alternatives}
 In order to verify the power of EPR knowledge, we have replaced the EPR knowledge with other expert knowledge and designed the following baselines, which are based on other expert knowledge as the alternatives of EPR knowledge to combine with GAIL:

\begin{itemize}
\item \textbf{{Distance}~\cite{deka2018factors}}: This baseline introduces in GAIL the knowledge that the probability of a user choosing a store for consumption is inversely proportional to the distance from the user to the store.

\item \textbf{{Price}}~\cite{pham2017factors}: This baseline combines GAIL with the knowledge that the probability of a user choosing a store for consumption is inversely proportional to the average price of the store.

\item \textbf{{Popularity}}~\cite{evangelista2019shopping}: This baseline introduces the knowledge that the probability of a user choosing a store for consumption is proportional to the historical average visit frequency of the store into GAIL.
\end{itemize}

We conducted experiments to compare EPR-GAIL with all the baselines on tasks of sales forecast and location recommendation on the Beijing and Guiyang data sets, as shown in the Table \ref{tab:alternatives_sales} and \ref{tab:alternatives_recommendation}. It can be seen from these two tables that EPR-GAIL based on EPR knowledge is at least 5\% better than baselines based on other expert knowledge, which verifies the power of EPR knowledge.

\subsection{Implementation Details}\label{Implementation Details}
In our experiments, we used Adam with learning rate of 3e-4, optimiser epsilon 1e-5, entropy coefficient 1e-3 for both actor and critic network of policy network. We used Adam with learning rate of 1e-4, optimiser epsilon 1e-5 for  reward  networks. The hy-perparameters of entropy, learning rate, network width (but not depth), and n-step returns have been optimised using a coarse grid search.  We train our model for about 1e+5 epochs using Pytorch 1.9 on $2\times2080Ti$ GPUs.
\newpage
\section{Global Response}

\subsection{Significance Analysis}

\begin{table*}[h]
    \centering
    \caption{Performance comparison and significance analysis over all baselines in two datasets. Each result is the average of five runs.}
    \resizebox{1.0\textwidth}{!}
    {
    \begin{tabular}{c||cccccc|cccccc}
         \Xhline{1pt} \textbf{Dataset}& \multicolumn{6}{c|}{\textbf{Beijing}}
        & \multicolumn{6}{c}{\textbf{Guiyang}}
        \\
        Metrics (JSD) &Consumption &Exploration &Price & Distance &Type &Identifier &Consumption &Exploration &Price & Distance&Type&Identifier  \\
        \hline \hline 
        IO-HMM&0.47&0.41&0.39&0.38&0.17&0.33&0.57&0.51&0.39&0.41&0.35&0.37\\
        EPR&0.13&0.12&0.48&0.21&0.38&0.27&0.14&0.15&0.47&0.26&0.34&0.35\\
        SeqGAN&0.39&0.28&0.31&0.32&0.29&0.35&0.41&0.38&0.36&0.45&0.33&0.37\\
        MoveSim&0.15&0.17&0.35&0.31&0.11&0.22&0.17&0.18&0.37&0.33&0.14&0.25\\
        \hline 
        BC&0.58&0.57&0.47&0.41&0.15&0.31&0.55&0.59&0.42&0.39&0.38&0.33\\
        GAIL&0.34&0.25&0.27&0.36&0.19&0.25&0.37&0.35&0.39&0.41&0.29&0.30\\
        \hline
        EPR-GAIL&\textbf{0.0004\textsuperscript{***}}&\textbf{0.0018\textsuperscript{***}}&\textbf{0.21\textsuperscript{**}}&\textbf{0.17\textsuperscript{**}}&\textbf{0.02\textsuperscript{***}}&\textbf{0.013\textsuperscript{***}}&\textbf{0.0008\textsuperscript{***}}&\textbf{0.0029\textsuperscript{***}}&\textbf{0.23\textsuperscript{**}}&\textbf{0.16\textsuperscript{**}}&\textbf{0.03\textsuperscript{***}}&\textbf{0.033\textsuperscript{***}}\\
        \hline
        Percentage&\textbf{96.92\%}&\textbf{98.50\%}&\textbf{22.22\%}&\textbf{19.04\%}&\textbf{81.81\%}&\textbf{94.09\%}&\textbf{99.42\%}&\textbf{98.07\%}&\textbf{36.11\%}&\textbf{38.46\%}&\textbf{78.57\%}&\textbf{86.80\%}\\
        \Xhline{1pt}
        \multicolumn{13}{l}{{\small  Note:\textsuperscript{*}p < 0.05, \textsuperscript{**}p < 0.01, \textsuperscript{***}p < 0.001.}}
    \end{tabular}
    }
\end{table*}

\subsection{Performance Comparison in the Purchase Setting with Higher Sparsity}

\begin{table*}[h]
    \centering
    \caption{Performance comparison in the purchase setting with higher sparsity in two datasets.}
    \resizebox{1.0\textwidth}{!}
    {
    \begin{tabular}{c||cccccc|cccccc}
         \Xhline{1pt} \textbf{Dataset}& \multicolumn{6}{c|}{\textbf{Beijing}}
        & \multicolumn{6}{c}{\textbf{Guiyang}}
        \\
        Metrics (JSD) &Consumption &Exploration &Price & Distance &Type &Identifier &Consumption &Exploration &Price & Distance&Type&Identifier  \\
        \hline \hline 
        MoveSim&0.35&0.34&0.49&0.42&0.31&0.32&0.37&0.39&0.51&0.43&0.33&0.34\\
        \hline
        EPR-GAIL&0.091&0.094&0.32&0.29&0.21&0.102&0.095&0.097&0.34&0.31&0.23&0.112\\
        \hline
        Percentage&\textbf{74.00\%}&\textbf{72.35\%}&\textbf{34.69\%}&\textbf{30.95\%}&\textbf{32.25\%}&\textbf{68.12\%}&\textbf{74.32\%}&\textbf{75.12\%}&\textbf{33.33\%}&\textbf{27.91\%}&\textbf{30.30\%}&\textbf{67.06\%}\\
        \Xhline{1pt}
    \end{tabular}
    }
\end{table*}

\subsection{Performance Comparison with CTRW and Levy flights}

\begin{table*}[h]
    \centering
    \caption{Performance comparison over all baselines in two datasets.}
    \resizebox{1.0\textwidth}{!}
    {
    \begin{tabular}{c||cccccc|cccccc}
         \Xhline{1pt} \textbf{Dataset}& \multicolumn{6}{c|}{\textbf{Beijing}}
        & \multicolumn{6}{c}{\textbf{Guiyang}}
        \\
        Metrics (JSD) &Consumption &Exploration &Price & Distance &Type &Identifier &Consumption &Exploration &Price & Distance&Type&Identifier  \\
        \hline \hline 
        CTRW&0.77&0.67&0.55&0.66&0.49&0.43&0.79&0.69&0.57&0.69&0.50&0.44\\
        Levy Flights&0.79&0.71&0.59&0.68&0.49&0.45&0.81&0.73&0.61&0.69&0.46&0.48\\
        IO-HMM&0.47&0.41&0.39&0.38&0.17&0.33&0.57&0.51&0.39&0.41&0.35&0.37\\
        EPR&0.13&0.12&0.48&0.21&0.38&0.27&0.14&0.15&0.47&0.26&0.34&0.35\\
        SeqGAN&0.39&0.28&0.31&0.32&0.29&0.35&0.41&0.38&0.36&0.45&0.33&0.37\\
        MoveSim&0.15&0.17&0.35&0.31&0.11&0.22&0.17&0.18&0.37&0.33&0.14&0.25\\
        \hline 
        BC&0.58&0.57&0.47&0.41&0.15&0.31&0.55&0.59&0.42&0.39&0.38&0.33\\
        GAIL&0.34&0.25&0.27&0.36&0.19&0.25&0.37&0.35&0.39&0.41&0.29&0.30\\
        \hline
        EPR-GAIL&\textbf{0.0004}&\textbf{0.0018}&\textbf{0.21}&\textbf{0.17}&\textbf{0.02}&\textbf{0.013}&\textbf{0.0008}&\textbf{0.0029}&\textbf{0.23}&\textbf{0.16}&\textbf{0.03}&\textbf{0.033}\\
        \hline
        Percentage&\textbf{96.92\%}&\textbf{98.50\%}&\textbf{22.22\%}&\textbf{19.04\%}&\textbf{81.81\%}&\textbf{94.09\%}&\textbf{99.42\%}&\textbf{98.07\%}&\textbf{36.11\%}&\textbf{38.46\%}&\textbf{78.57\%}&\textbf{86.80\%}\\
        \Xhline{1pt}
    \end{tabular}
    }
\end{table*} 
\end{document}